\documentclass[letterpaper]{article} 
\usepackage{aaai2026}  
\usepackage{times}  
\usepackage{helvet}  
\usepackage{courier}  
\usepackage[hyphens]{url}  
\usepackage{graphicx} 
\usepackage{natbib}  
\usepackage{caption} 
\usepackage{algorithm}
\usepackage{algorithmic}
\usepackage{amsmath}
\usepackage{amssymb}
\usepackage{booktabs}
\usepackage{multirow}
\usepackage{xcolor}

\usepackage{newfloat}
\usepackage{listings}
\DeclareCaptionStyle{ruled}{labelfont=normalfont,labelsep=colon,strut=off} 
\floatstyle{ruled}
\newfloat{listing}{tb}{lst}{}
\floatname{listing}{Listing}
\nocopyright 

\title{DynaMind: Reconstructing Dynamic Visual Scenes from EEG by Aligning Temporal Dynamics and Multimodal Semantics to Guided Diffusion}

\author {
    Junxiang Liu\equalcontrib,
    Junming Lin\equalcontrib,
    Jiangtong Li,
    Jie Li
}
\affiliations {
    Tongji University\\
    \{junxiang\_liu,jiangtongli\}@tongji.edu.cn
}

\usepackage{bibentry}

\begin{document}

\maketitle

\begin{abstract}
Reconstruction dynamic visual scenes from electroencephalography~(EEG) signals remains a primary challenge in brain decoding, limited by the low spatial resolution of EEG, a temporal mismatch between neural recordings and video dynamics, and the insufficient use of semantic information within brain activity.
Therefore, existing methods often inadequately resolve both the dynamic coherence and the complex semantic context of the perceived visual stimuli.
To overcome these limitations, we introduce \textbf{DynaMind}, a novel framework that reconstructs video by jointly modeling neural dynamics and semantic features via three core modules: a Regional-aware Semantic Mapper~(\textbf{RSM}), a Temporal-aware Dynamic Aligner~(\textbf{TDA}), and a Dual-Guidance Video Reconstructor (\textbf{DGVR}).
The \textbf{RSM} first utilizes a regional-aware encoder to extract multimodal semantic features from EEG signals across distinct brain regions, aggregating them into a unified diffusion prior.
In the mean time, the \textbf{TDA} generates a dynamic latent sequence, or blueprint, to enforce temporal consistency between the feature representations and the original neural recordings.
Together, guided by the semantic diffusion prior, the \textbf{DGVR} translates the temporal-aware blueprint into a high-fidelity video reconstruction.
On the SEED-DV dataset, \textbf{DynaMind} sets a new state-of-the-art (SOTA), boosting reconstructed video accuracies (video- and frame-based) by 12.5 and 10.3 percentage points, respectively. 
It also achieves a leap in pixel-level quality, showing exceptional visual fidelity and temporal coherence with a 9.4\% SSIM improvement and a 19.7\% FVMD reduction. 
This marks a critical advancement, bridging the gap between neural dynamics and high-fidelity visual semantics.
\end{abstract}

\section{Introduction}

EEG signals, which contain rich human cognitive information~\cite{dahal2011modeling}, have been successfully applied to tasks such as motor imagery~\cite{al2021deep} and emotion recognition~\cite{liu2020subject}.
As visual stimuli constitute over 60\% of sensory input~\cite{scholler2012toward}, extracting visual information from EEG is crucial for understanding fundamental cognitive processes~\cite{song2023decoding}.
Pioneering research, such as~\citet{singh2023eeg2image} and~\citet{wei2024mb2c}, has advanced this area, establishing visual reconstruction from EEG as a significant emerging field.
Contemporary methods leverage powerful deep learning models such as CLIP~\cite{radford2021learning} and Stable Diffusion~(SD)~\cite{rombach2022high}.
The prevailing framework projects EEG features into the latent space of CLIP to derive semantic representations of the visual stimuli, which then condition generative models for image synthesis.
The success of this approach in achieving high image quality and semantic accuracy has established a reliable and efficient generative paradigm.
\textit{This success naturally paves the way for the next frontier in EEG based reconstruction, \emph{i.e.}, capturing the dynamic visual world from brain activity to generate coherent video sequences.}

\begin{figure}[t]
 \centering
 \includegraphics[width=\columnwidth]{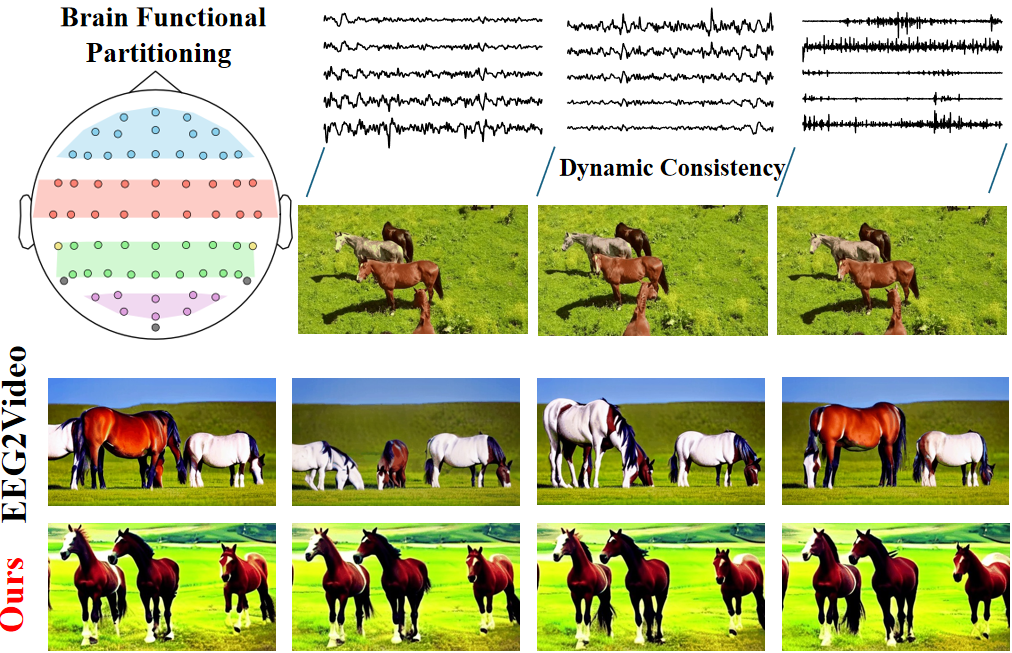} 
 \caption{Our approach captures spatio features from diverse brain regions and enforce temporal dynamic consistency between EEG signals and video. This results in generated videos with superior fidelity, temporal coherence, and semantic accuracy compared to EEG2Video.}
 \label{fig:comparison}
\end{figure}

The domain of video reconstruction from EEG, while nascent, presents significant scientific and technical challenges.
As illustrated in Fig.~\ref{fig:comparison}, distinct functional brain regions exhibit differential patterns of activity when processing complex visual information~\cite{li2024visual}.
Furthermore, neural responses to dynamic visual cues, such as motion velocity and direction, are encoded by distinct cortical areas~\cite{goodale1992separate}.
However, current research predominantly extracts semantic information from the occipital lobe, often neglecting valuable cognitive information from other brain areas~\cite{zhang2025cognitioncapturer}.
Therefore, by attempting to extract only visual information from EEG~\cite{liu2024eeg2text}, these approaches overlook the value of critical multimodal information within the EEG signals themselves.
Moreover, existing studies lack dynamic consistency between stimulus representation and the original neural recordings.
As shown in Fig.~\ref{fig:comparison}, models like EEG2Video~\cite{liu2024eeg2video} often produce reconstructions with intra-frame semantic errors and inter-frame temporal inconsistencies.
\textit{Collectively, these limitations present a significant barrier for high-fidelity EEG-to-Video reconstruction.}

To address these challenges, we propose \textbf{DynaMind}, a novel framework designed to enrich semantic features using a brain-functional partitioning approach while enforcing dynamic coherence with the video content.
The framework accomplishes this task through the synergistic operation of three key components: a Regional-aware Semantic Mapper~(\textbf{RSM}), a Temporal-aware Dynamic Aligner~(\textbf{TDA}), and a Dual-Guidance Video Reconstructor~(\textbf{DGVR}).
\textbf{RSM} extracts features from distinct functional brain regions, employs multimodal constraints to enrich their semantic diversity, and then aggregates these features into a unified diffusion prior.
Meanwhile, \textbf{TDA} addresses temporal dynamics by exploiting EEG features to generate a blueprint sequence that enforces dynamic consistency between the neural recordings and video stimulus.
Building upon this foundation, \textbf{DGVR} leverages the multimodal diffusion prior as semantic guidance to translate the dynamically coherent temporal blueprint into high-fidelity videos exhibiting both temporal consistency and semantic accuracy.

Experiments on the SEED-DV dataset~\cite{liu2024eeg2video} demonstrate that our \textbf{DynaMind} framework not only improves direct EEG classification accuracy and the classification accuracy based on its reconstructed videos, but also achieves a leap in pixel-level quality. 
This is evidenced by a 9.4\% improvement in frame-level Structural Similarity (SSIM) and a 19.7\% reduction in Fréchet Video Motion Distance (FVMD).
Furthermore, we conduct detailed ablation studies to analyze the contributions of different brain regions and the supportive role of multimodal feature integration
Overall, our results mark a significant advancement in the decoding of dynamic visual perception from EEG signals.
Our main contributions are summarized as follows:
\begin{itemize}
    \item We identify and analyze key limitations in EEG-based video reconstruction, namely the underutilization of regional brain information and the poor temporal consistency of existing methods.
    \item We propose DynaMind, a novel three-module framework designed to improve the dynamic coherence and semantic richness of video reconstructions from EEG signals.
    \item We validate our framework on the public SEED-DV dataset, establishing a new SOTA in both classification accuracy and reconstruction quality, thus providing a new benchmark for the field.
\end{itemize}

\section{Related works}

\subsection{Decoding Visual Information from Brain}

A substantial body of research has focused on decoding visual information from physiological signals, utilizing modalities such as functional magnetic resonance imaging~(fMRI)~\cite{du2018reconstructing}, magnetoencephalography~(MEG)~\cite{adjamian2008induced}, and electroencephalography~(EEG)~\cite{sun2023contrast}. 
To accomplish this, a variety of generative models have been explored, prominently featuring variational autoencoders~(VAEs)~\cite{kavasidis2017brain2image} and generative adversarial networks~(GANs)~\cite{song2023decoding}. 
More recently, the advent of advanced diffusion models like Stable Diffusion~(SD) has spurred significant progress in visual reconstruction~\cite{rombach2022high}. 
Generative schemes employing these models primarily rely on rich embeddings to guide the diffusion process toward specific semantic objectives. 
Therefore, numerous studies have attempted to derive such semantic information from EEG~\cite{song2023decoding,liu2024eeg2text}. 
Its inherent high temporal resolution presents a unique and critical advantage for the more complex task of dynamic video reconstruction.

However, the high signal-to-noise and low spatial resolution ratio of EEG still renders feature extraction difficult. 
Current methods often attempt to solve this by focusing exclusively on the relationship between EEG and image semantics, overlooking valuable multimodal information inherent in the brain signals themselves. 
This limitation becomes particularly salient in EEG-to-video generation, where it may lead to videos that are conceptually approximate but exhibit deviations. 
While pioneering work such as EEG2Video~\cite{liu2024eeg2video} has ventured into this area, its resulting limitations in semantic accuracy and temporal coherence highlight the critical need for further research in the field of video reconstruction from brain signals.

\subsection{Video Generation Models}

Breakthroughs in Text-to-Image~(T2I) generation have been largely driven by large-scale multimodal datasets composed of billions of text-image pairs~\cite{nichol2021glide, saharia2022photorealistic}.
To replicate this success in Text-to-Video~(T2V) generation, recent studies~\cite{singer2022make, villegas2022phenaki} have extended these space-only T2I models to the spatio-temporal domain.
By training on large-scale text-video datasets such as WebVid-10M~\cite{bain2021frozen}, these models have achieved promising results, laying a foundation for reconstructing visuals from other modalities.

However, prior research in brain signal-based reconstruction has typically relied on holistic semantics from the entire signal to directly guide the generative process.
Fundamentally, successful video generation hinges on preserving the continuous motion of consistent objects across frames~\cite{wu2023tune}.
As shown in the 2nd row of Fig.~\ref{fig:comparison}, this reliance on holistic guidance leads to reconstructions with significant temporal inconsistencies between consecutive frames.

\section{Methodology}

\begin{figure*}[h]
 \centering
 \includegraphics[width=0.95\textwidth]{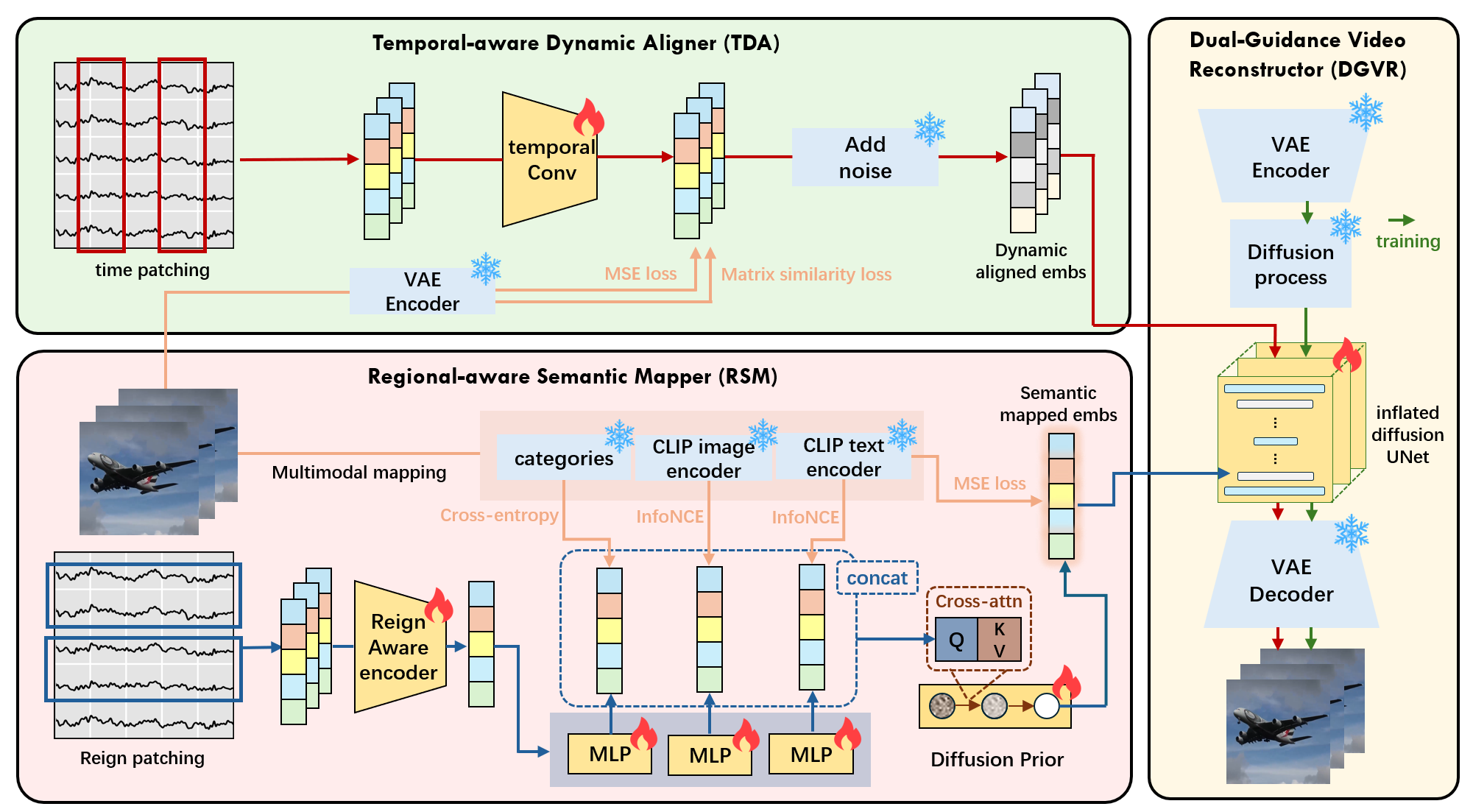} 
 \caption{An overview of our DynaMind framework: A dual-guidance architecture for reconstructing high-fidelity videos from EEG signals.
 Regional-aware Semantic Mapper~(RSM) processes EEG signals by brain regions to extract a semantically rich prior, capturing the multimodal content of visual experience. 
 In parallel, Temporal-aware Dynamic Aligner~(TDA) processes the global EEG sequence to generate a dynamic temporal blueprint, governing how visual scene evolves over time.
 Finally, Dual-Guidance Video Reconstructor~(DGVR), a latent diffusion model, utilizes the semantic prior for conditional guidance and the temporal blueprint as initial latent variables to synthesize videos, ensuring both semantic accuracy and temporal coherence.}
 \label{fig:Framework}
\end{figure*}

As shown in Figure~\ref{fig:Framework}, our DynaMind framework employs three synergistic modules to translate raw EEG signals into high-fidelity video reconstructions.
Regional-aware Semantic Mapper~(RSM) generates a semantic diffusion prior by interpreting distinct patterns of regional brain activity. Temporal-aware Dynamic Aligner~(TDA) produces a coherent temporal blueprint to capture the corresponding visual dynamics. Together, Dual-Guidance Video Reconstructor~(DGVR), from a pre-trained video diffusion model, is optimized to accept dual guidance from both the semantic prior and the temporal blueprint, enabling the generated videos with high semantic accuracy and temporal coherence.

\subsection{Regional-aware Semantic Mapper}
\label{RSM}
\subsubsection{Region-Aware EEG Representation Learning}
Human cognitive processes are distributed across functionally specialized networks in the brain.
For instance, the occipital lobe is primarily involved in processing visual stimuli, while the temporal and frontal lobes manage more abstract semantic and emotional contexts~\cite{mesulam1998sensation}.
Therefore, \textbf{RSM} is designed to capture this diversity, leveraging the functional specialization of brain regions to extract semantically rich and varied information from EEG signals.

The process begins by partitioning the multi-channel EEG signals $\mathbf{E} \in \mathbb{R}^{C \times T}$, where $C$ is the channel count and $T$ is the time points, into $K$ distinct groups $\{\mathbf{E}_1, \mathbf{E}_2, \cdots, \mathbf{E}_K\}$ based on anatomical brain regions (\emph{i.e.}, frontal, central, temporal, parietal, and occipital lobes).
Each regional signal group $\mathbf{E}_i$ is then processed by a dedicated region-aware encoder. 
This encoder, consisting of temporal convolutions and a Transformer layer, captures region-specific neural patterns to produce a feature embedding $f_i$. 
These regional embeddings $\{\mathbf{f}_1, \mathbf{f}_2, \cdots, \mathbf{f}_K\}$ are fused together to $\mathbf{F}$.
\begin{align}
    \mathbf{f} = \text{Concat}(\mathbf{f}_1, \mathbf{f}_2, \cdots, \mathbf{f}_K),\ \  \text{where} \ \ 
    \mathbf{f}_i = \text{Enc}_i(\mathbf{E}_i),
\end{align}
where each $\text{Enc}_i$ has independent parameters, allowing it to specialize in its designated brain region's signals.

\subsubsection{Semantic Enrichment via Multi-Modal Alignment}
To enrich these fused embeddings with diverse semantics, we project them into multiple latent spaces for alignment with corresponding multimodal CLIP embeddings.
Specifically, we employ a set of mapping networks, denoted as $M_i$, to transform the fused EEG representation $\mathbf{f}$ into the latent spaces of the target multimodal embeddings.
\begin{equation}
    \mathbf{\hat{c}_I} = M_{\phi_{I}}(\mathbf{f}); \,\mathbf{\hat{c}_T} = M_{\phi_{T}}(\mathbf{f}); \,\mathbf{\hat{c}_Y} = M_{\phi_{Y}}(\mathbf{f}),
\end{equation}
where $\phi_I, \phi_T, \text{and } \phi_Y$ are parameters for each mapping network, which are implemented as multi-layer perceptrons.

The alignment between the predicted embeddings ($\mathbf{\hat{c}}$) and ground-truth embeddings ($\mathbf{c}$) is optimized using the InfoNCE loss~\cite{oord2018representation}, defined as
\begin{align}
    \mathcal{L}_{\mathbf{\hat{c}} \to \mathbf{c}} &= -\log \frac{\exp(\text{sim}(\mathbf{\hat{c}}, \mathbf{c}) / \tau)}{\sum_{j=1}^{B} \exp(\text{sim}(\mathbf{\hat{c}}, \mathbf{c_{j}}) / \tau)}, \\
    \mathcal{L}_{\mathbf{c} \to \mathbf{\hat{c}}} &= -\log \frac{\exp(\text{sim}(\mathbf{c}, \mathbf{\hat{c}}) / \tau)}{\sum_{j=1}^{B} \exp(\text{sim}(\mathbf{c}, \mathbf{\hat{c_{j}}}) / \tau)}, \\
    \mathcal{L}_{\text{info}}(\mathbf{\hat{c}}, \mathbf{{c}}) &= \frac{1}{2}(\mathcal{L}_{\mathbf{\hat{c}} \to \mathbf{c}} + \mathcal{L}_{\mathbf{c} \to \mathbf{\hat{c}}}).\label{equ:infonce}
\end{align}

where $\text{sim}(\cdot, \cdot)$ is cosine similarity and $\tau$ is a temperature hyperparameter. 
This symmetric formulation robustly aligns the two embedding spaces. 
Additionally, a cross-entropy (CE) classification loss is applied to the predicted embedding $\mathbf{\hat{c}}_Y$ after a linear projection to class logits $\mathbf{\hat{y}}$:
\begin{align}
\mathcal{L}_{\text{category}} = -\frac{1}{N} \sum_{i=1}^N \left( \mathbf{y}_i \cdot \log(\text{Softmax}(\hat{\mathbf{y}}_i)) \right).
\end{align}
where $\hat{\mathbf{y}}$ and $\mathbf{y}$ represent the predicted class probabilities and one-hot ground-truth labels, respectively.

\subsubsection{Generation of Semantic Diffusion Prior}
The generation of high-fidelity video content hinges on producing effective latent embeddings to condition the subsequent generative model.
We observe that embeddings within the CLIP text space provide more effective conditioning for the reconstruction process than the direct EEG-derived features.
Therefore, translating the EEG-derived features into the CLIP text embedding space is a critical step for enhancing the final reconstruction quality.
Inspired by DALLE-2~\cite{ramesh2022hierarchical}, we employ a diffusion prior to perform this cross-modal translation.
Specifically, this diffusion prior model learns to map the EEG embedding $\mathbf{\hat{c}}_{\text{diff}}$, to a corresponding target embedding within the CLIP text space.
For training this prior model, we utilize the same prior loss, $L_{prior}$, as defined in DALLE-2~\cite{ramesh2022hierarchical}.

\begin{align} 
\mathbf{\hat{c}}_{\text{diff}} &= \text{Concat}(\mathbf{\hat{c}}_I, \mathbf{\hat{c}}_T, \mathbf{\hat{c}}_Y), \\
\mathcal{L}_{\text{prior}} &= \mathbb{E}_{t \sim [1, T] } \left[ \left\| P_\theta(\mathbf{\hat{c}_T}^{(t)}, t, \mathbf{\hat{c}}_{\text{diff}}) - \mathbf{\hat{c}_T} \right\|^2 \right].
\end{align}
The total loss is a summation of the InfoNCE losses from image and text, the classification loss and the diffusion loss:
\begin{equation}
    \mathcal{L}_{\text{align}} =  \mathcal{L}_{\text{info}}(\mathbf{\hat{c}_{I}}, \mathbf{{c}_{I}}) +  \mathcal{L}_{\text{info}}(\mathbf{\hat{c}_{T}}, \mathbf{{c}_{T}}) +  \mathcal{L}_{\text{category}} +\mathcal{L}_{\text{prior}}.
\end{equation}

\subsection{Temporal-aware Dynamic Aligner}
Unlike the RSM, which focuses on semantic content, the TDA is specifically designed to address the critical challenge of temporal coherence.
This module generates a dynamic temporal blueprint to ensure that the motion and flow of the reconstructed video remain synchronized with the underlying temporal dynamics of the EEG signals.
Therefore, TDA is vital for mitigating the inter-frame inconsistencies that often plague prior EEG-to-video synthesis methods.
To achieve this, the TDA operates in two main stages: the direct generation of a temporal blueprint and its subsequent contrastive alignment with ground-truth video features.

\subsubsection{Extraction of Temporal Blueprint}
The \textbf{TDA} receives the raw EEG signal sequence $E \in \mathbb{R}^{C \times T}$ as input. This sequence is first segmented along the temporal dimension into $N$ non-overlapping windows, $\{\mathbf{E}^{t}_1, \mathbf{E}^{t}_2, \cdots, \mathbf{E}^{t}_N\}$, where each window $\mathbf{E}^{t}_i \in \mathbb{R}^{C \times \frac{T}{N}}$.
Each temporal segment $\mathbf{E}^{t}_i$ is then processed by temporal convolutional network, denoted $\text{TCN}_{temp}$. This architecture is selected for its proficiency in modeling long-range dependencies and is optimized here to capture rhythmic and transitional patterns within the neural signals.
The concatenated outputs from the TCN across all segments form the \textbf{dynamic temporal blueprint}, $\mathbf{H}_{temp}$:
\begin{align}
  \mathbf{H}_{temp} &= [\mathbf{h_1}, \mathbf{h_2}, \ldots, \mathbf{h_N}] \\
  \mathbf{h}_{i} &= \text{TCN}_{temp}(\mathbf{E}^{t}_i).
\end{align}
This dynamic temporal blueprint is a sequence of latent vectors, where $\mathbf{h}_i$ encodes the anticipated motion and state changes for the corresponding $i$-th segment of the video.

\subsubsection{Contrastive Alignment with Video Features}

To ensure the temporal blueprint $\mathbf{H}_{\text{temp}}$ captures both the content and temporal dynamics of the corresponding video, we align it with video features using a dual objective that enforces both content-wise and structural consistency.
First, we define the input features. The temporal blueprint from EEG signal is a feature sequence $\mathbf{H}_{\text{temp}} \in \mathbb{R}^{N \times D_{\text{eeg}}}$, where $N$ is the number of frames. 
The corresponding video is processed by a VAE encoder and a flattening layer to yield a sequence of frame-level visual features $\mathbf{V} \in \mathbb{R}^{N \times D_{\text{vid}}}$.
Next, both sequences are projected into a shared latent space of dimension $D_{\text{latent}}$ using distinct projection heads, $P_H$ and $P_V$:
\begin{align}
    \mathbf{Z}_H &= P_H(\mathbf{H}_{\text{temp}}) \in \mathbb{R}^{N \times D_{\text{latent}}},\\ \mathbf{Z}_V &= P_V(\mathbf{V}') \in \mathbb{R}^{N \times D_{\text{latent}}}.
\end{align}
With the features projected into this common space, we apply two alignment losses. To enforce direct, frame-by-frame content similarity, we compute the Mean Squared Error~(MSE) between the two projected sequences:
\begin{equation} 
\mathcal{L}_{\text{HV}} = \frac{1}{N} \sum_{i=1}^{N} \left\| (\mathbf{Z}_H)_i - (\mathbf{Z}_V)_i \right\|^2.
\end{equation}
Moreover, to align the relational dynamics between frames, we introduce a structural loss. 
This loss compares the intra-modality similarity matrices, $\mathbf{S}_H \in \mathbb{R}^{N \times N}$ and $\mathbf{S}_V \in \mathbb{R}^{N \times N}$, which capture the internal structure of each sequence. 
The matrices are computed using cosine similarity, \textit{e.g.}, $(\mathbf{S}_H)_{ij} = \text{cos-sim}((\mathbf{Z}_H)_i, (\mathbf{Z}_H)_j)$. 
The loss then minimizes the MSE between these two structural representations:
\begin{equation} 
\mathcal{L}_{\text{Struct}} = \frac{1}{N^2} \left\| \mathbf{S}_H - \mathbf{S}_V \right\|_F^2.
\end{equation}
The final loss for the TDA module is the sum of the content and structural alignment losses:
\begin{equation}
\mathcal{L}_{\text{TDA}} = \mathcal{L}_{\text{HV}} + \mathcal{L}_{\text{Struct}}.
\end{equation}

\begin{table*}[t]
\centering
\resizebox{\textwidth}{!}{%
\setlength{\tabcolsep}{1mm} 
\begin{tabular}{lcccccccccc}
\toprule
\textbf{Method} & \textbf{40-c top-1} & \textbf{40-c top-5} & \textbf{9-c top-1} & \textbf{9-c top-3} & \textbf{Color} & \textbf{Fast/Slow} & \textbf{Numbers} & \textbf{Human Face} & \textbf{Human} \\
\midrule
Chance level & 2.50 & 12.50 & 11.11 & 33.33 & 20.57 & 50.00 & 65.64 & 62.25 & 71.43 \\
\midrule
ShallowNet & 5.59/2.27* & 16.93/4.66* & 21.40/1.96* & 49.62/2.34* & 27.00/2.09* & 56.62/1.77* & 66.15/0.89 & 64.87/1.54 & 73.21/1.52 \\
DeepNet & 4.56/1.52* & 14.30/3.25* & 20.27/1.25* & 48.06/1.59* & 26.37/1.95* & 55.42/0.59* & 65.71/0.24 & 61.58/3.93 & 72.86/0.40 \\
EEGNet & 4.64/0.86* & 14.25/1.87* & 19.63/0.81* & 47.04/1.45* & 25.46/1.31* & 51.99/2.00 & 64.67/0.60 & 61.37/1.31 & 72.38/0.98 \\
Conformer & 4.93/1.57* & 15.36/4.44* & 20.92/0.98* & 49.25/1.49* & \textbf{27.53/1.37*} & 55.02/0.83* & 65.73/0.26 & 64.96/1.14 & 73.00/0.85 \\
TSConv & 4.92/0.99* & 15.05/2.31* & 20.00/1.01* & 47.76/1.51* & 26.89/1.83* & 55.32/0.99* & 65.39/0.41 & 64.39/1.47 & 72.68/0.67 \\
GLMNet & 6.20/3.02* & 17.75/4.24* & 21.93/1.87* & 50.01/2.52* & 27.33/1.45* & \textbf{57.35/1.98*} & 66.21/0.91 & 65.10/1.45 & 73.34/1.31 \\  
\textbf{Ours} & \textbf{8.27/2.85*} & \textbf{22.38/4.80*} & \textbf{22.73/3.07*} & \textbf{51.93/4.73*} & 27.43/1.11* & 56.44/1.51* & \textbf{67.56/0.94} & \textbf{82.47/0.81*} & \textbf{73.64/1.70} \\
\bottomrule

\end{tabular}
}
\caption{Average classification accuracy (\%) and std across all subjects with different EEG classifiers on different tasks. Chance level is the percentage of the largest class. The star symbol (*) represents the result is above chance level with statistical significance (two-sample t-test: $p < 0.05$).}
\label{tab:classification_comparison}
\end{table*}

\begin{table*}[htbp]
\centering
\begin{tabular}{llcccccc}
\toprule
&\multirow{2}{*}{Metrics} & \multicolumn{3}{c}{\textbf{Video-based}} & \multicolumn{3}{c}{\textbf{Frame-based}} \\
\cmidrule(lr){3-5} \cmidrule(lr){6-8}
& & \multicolumn{2}{c}{Semantic-level} & Pixel-level & \multicolumn{2}{c}{Semantic-level} & Pixel-level \\
\cmidrule(lr){3-4} \cmidrule(lr){6-7}
\textbf{\# Classes} & \textbf{Method} & 2-way & 40-way & FVMD$\downarrow$ & 2-way & 40-way & SSIM$\uparrow$ \\
\midrule
\multirow{2}{*}{10} & Ours & 0.847$\pm$0.01 & 0.394$\pm$0.03 & 1601.84$\pm$0.02 & 0.833$\pm$0.02 & 0.308$\pm$0.01 & 0.309$\pm$0.02 \\
& EEG2Video & 0.852$\pm$0.02 & 0.340$\pm$0.01 & 1977.13$\pm$0.01 & 0.798$\pm$0.03 & 0.232$\pm$0.02 & 0.300$\pm$0.03 \\ 
\midrule
\multirow{2}{*}{20} & Ours & 0.835$\pm$0.01 & 0.345$\pm$0.01 & 1587.91$\pm$0.01 & 0.818$\pm$0.02 & 0.277$\pm$0.02 & 0.290$\pm$0.02 \\
& EEG2Video &  0.813$\pm$0.02 & 0.273$\pm$0.03 & 1960.20$\pm$0.03 & 0.785$\pm$0.04 & 0.184$\pm$0.02 
& 0.242$\pm$0.03 \\
\midrule
\multirow{2}{*}{30} & Ours & 0.833$\pm$0.02 & 0.309$\pm$0.01 & 1661.06$\pm$0.03 & 0.805$\pm$0.02 & 0.254$\pm$0.01 & 0.293$\pm$0.01 \\
& EEG2Video & 0.794$\pm$0.02 & 0.209$\pm$0.05 & 2016.67$\pm$0.04 & 0.785$\pm$0.04 & 0.180$\pm$0.02 & 0.228$\pm$0.04 \\
\midrule
\multirow{2}{*}{40} & Ours & 0.828$\pm$0.02 & 0.284$\pm$0.02 & 1637.55$\pm$0.01 & 0.807$\pm$0.03 & 0.241$\pm$0.01 & 0.280$\pm$0.01 \\
& EEG2Video & 0.798$\pm$0.03 &  0.159$\pm$0.01 & 2038.27$\pm$0.02 & 0.774$\pm$0.02 & 0.138$\pm$0.01 & 0.256$\pm$0.03 \\
\bottomrule
\end{tabular}
\caption{Reconstruction results across varying class numbers, evaluated on semantic/pixel metric at the video/frame level.}
\label{tab:reconstruction_comparison}
\end{table*}

\subsection{Dual-Guidance Video Reconstructor (DGVR)}
The \textbf{DGVR} is responsible for synthesizing the final high-fidelity video. 
Adopting the Tune-A-Video technique~\cite{wu2023tune} as a foundation, our model is conditioned on both the semantic diffusion prior from the \textbf{RSM} and the temporal blueprint from the \textbf{TDA}. 
This dual-guidance mechanism is key to generating videos that are both semantically accurate and temporally coherent.

Instead of initiating the reverse diffusion process from pure Gaussian noise, we structure the initial latent variable $\mathbf{x}_T$ using the temporal blueprint $\mathbf{H}_{temp}$. 
This ensures that the foundational spatio-temporal structure of the video is aligned with the neural dynamics from the outset. 
Specifically, we first project the blueprint to a latent representation $\mathbf{z}_{B}$ and define the initial latent as:
\begin{equation}
    \mathbf{x}_T = \mathcal{E} + \alpha \cdot \mathcal{U}(\mathbf{z}_{B}),
\end{equation}
where $\mathcal{E} \sim \mathcal{N}(0, \mathbf{I})$ is a random Gaussian noise tensor, $\mathcal{U}(\cdot)$ is an upsampling network matching the blueprint's dimensions to the latent space, and $\alpha$ is a hyperparameter controlling the temporal guidance strength.
During each step of the reverse diffusion process (from $t = T$ down to 1), the semantic diffusion prior, $\mathbf{\hat{c}}_{diff}$, is injected into the model's U-Net architecture via cross-attention layers.
This conditioning ensures that the objects, scenes, and overall context of the generated frames are semantically consistent with the information decoded from the various brain regions.

The diffusion model's U-Net, denoted $\epsilon_\theta$, is trained to predict the noise $\epsilon$ from a noisy latent $\mathbf{x}_t$ at a given timestep $t$. The prediction is conditioned on $\mathbf{x}_t$, $t$, the semantic diffusion prior, $\mathbf{\hat{c}}_\text{diff}$, and the projected temporal blueprint $\mathbf{z}_{B}$. The objective function is the standard diffusion loss:
\begin{equation} 
\mathcal{L}_{\text{DGVR}} = \mathbb{E}_{\mathbf{x}_0, \epsilon, t} \left[ \left\| \epsilon - \epsilon_\theta (\mathbf{x}_t, t, \mathbf{\hat{c}}_{\text{diff}}, \mathbf{z}_{B}) \right\|^2 \right].
\end{equation}
where $\mathbf{x}_t = \sqrt{\bar{\alpha}_t}\mathbf{x}_0 + \sqrt{1 - \bar{\alpha}_t}\epsilon$, and $\epsilon \sim \mathcal{N}(0, \mathbf{I})$.

After the iterative denoising process is complete, a final decoder module transforms the clean latent representation $\mathbf{x}_0$ into the pixel-space video. By leveraging both dynamic and semantic guidance, our model reconstructs videos with superior fidelity, coherence, and contextual accuracy.

\begin{table*}[htbp]
\centering
\begin{tabular}{lccccccccc}
\toprule
& & \multicolumn{2}{c}{\textbf{Classification-based}}& \multicolumn{3}{c}{\textbf{Video-based}} & \multicolumn{3}{c}{\textbf{Frame-based}} \\
\cmidrule(lr){3-4} \cmidrule(lr){5-7} \cmidrule(lr){8-10}
&\multirow{2}{*}{\textbf{Method}} & \multirow{2}{*}{40-c top-1}  & \multirow{2}{*}{9-c top-1} & \multicolumn{2}{c}{Semantic-level} & Pixel-level & \multicolumn{2}{c}{Semantic-level} & Pixel-level \\
\cmidrule(lr){5-6} \cmidrule(lr){8-9}
&  & & & 2-way & 40-way & FVMD$\downarrow$ & 2-way & 40-way & SSIM$\uparrow$ \\
\midrule
 & \textbf{Ours} & \textbf{8.27} & \textbf{22.73} & \textbf{0.828} & \textbf{0.284} & 1637.55 & 0.807 & \textbf{0.241} & \textbf{0.280} \\ 
\midrule
\multicolumn{10}{c}{\textit{Brain Regions}} \\
\midrule
& w/o Frontal & 7.67 & 22.02 & 0.813 & 0.264 & 1677.42 & 0.793 & 0.220 & 0.267 \\ 
& w/o Parietal & 7.65 & 22.01 & 0.812 & 0.263 & \textbf{1631.21} & 0.796 & 0.221 & 0.267 \\ 
& w/o Occipital & 6.73 & 20.93 & 0.807 & 0.239 & 1712.11 & 0.786 & 0.189 & 0.253 \\ 
& w/o Temporal & 7.22 & 21.57 & 0.809 & 0.252 & 1658.89 & 0.790 & 0.205 & 0.260 \\ 
\midrule
\multicolumn{10}{c}{\textit{Features}} \\
\midrule
& w/o Image & 8.08 & 22.44 & 0.825 & 0.281 & 1649.77 & 0.807 & 0.235 & 0.277 \\
& w/o Text & 7.56 & 21.87 & 0.811 & 0.252 & 1703.48 & 0.793 & 0.212 & 0.270 \\
& w/o Category & 8.11 & 22.45 & 0.820 & 0.276 & 1644.07 & 0.802 & 0.232 & 0.278 \\
\midrule
\multicolumn{10}{c}{\textit{Consistency}} \\
\midrule
& w/o Consistency & - & - & 0.825 & 0.279 & 1916.91 & \textbf{0.810} & 0.240 & \textbf{0.280} \\
\bottomrule
\end{tabular}
\caption{Quantitative results of the ablation study on the key components of our model. The performance drop after removing individual brain regions, features, or the consistency module validates the effectiveness of each component.}
\label{tab:ablation study}
\vspace{-2pt}
\end{table*}

\section{Experiments}
\subsection{Experimental Setup}

\noindent\textbf{Dataset and Baselines.}
Our experiments are conducted on the \textbf{SEED-DV} dataset~\cite{liu2024eeg2video}, which comprises EEG signals from 20 subjects viewing video clips across 40 visual conceptual categories.
To validate the effectiveness of our model, we benchmark its performance against comprehensive baseline methods, including both foundational EEG models~(ShallowNet~\cite{schirrmeister2017deep} and EEGNet~\cite{lawhern2018eegnet}), and SOTA approaches~(Conformer~\cite{song2022eeg} and GLMNet~\cite{liu2024eeg2video}).

\noindent\textbf{Implementation and Evaluation.}
Our model is composed of three core modules. 
The RSM aligns regional EEG embeddings with semantic features from a pre-trained CLIP ViT-L/14 model. 
In parallel, the TDA extracts a temporal blueprint from the global EEG dynamics. 
Finally, the DGVR fine-tunes a Stable Diffusion v1.4 model to synthesize 6-frame video clips at 512 $\times$ 288 resolution (3 fps), conditioned on the semantic and temporal guidance from the other modules.
We follow the evaluation protocol established by EEG2Video~\cite{liu2024eeg2video} to ensure fair comparison. 
Our assessment includes 
1) Frame-Level Quality: The Structural Similarity Index (SSIM)~\cite{wang2004image} is used to measure pixel-level fidelity. 
N-way top-K classification accuracy based on CLIP features is used to assess semantic correctness;
2) Video-Level Dynamics: N-way top-K accuracy based on VideoMAE features is used to evaluate dynamic semantic coherence. 
The Frame-wise Video Motion Distance (FVMD)~\cite{liu2024fr} is used to quantify temporal consistency.
\textbf{More details about the implementation and evaluation are in Appendix.}

\begin{figure*}[h]
 \centering
 \includegraphics[width=0.98\textwidth]{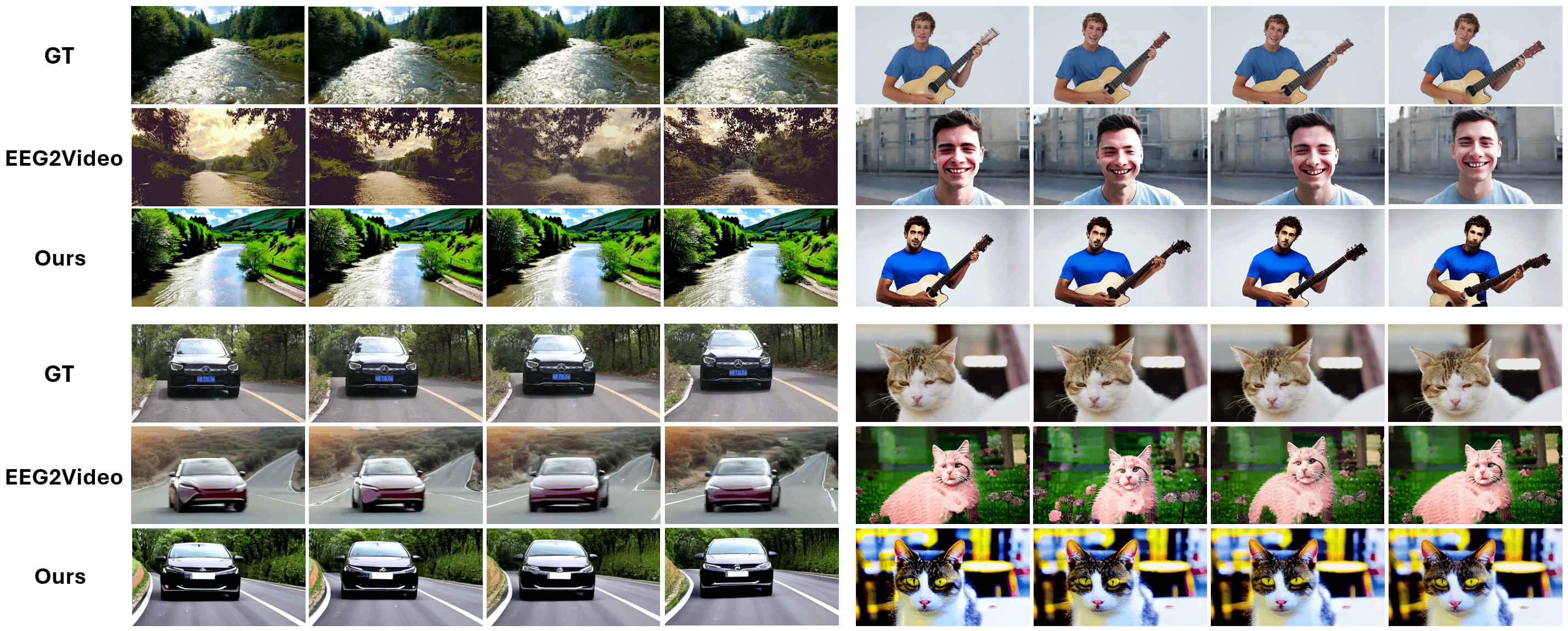} 
 \caption{Video reconstruction results of ours and EEG2Video compared against the ground truth for diverse visual concepts. The results of EEG2Video are from their openly released samples.}
 \label{fig:Visualizaion}
\end{figure*}

\subsection{Main Results}
\noindent\textbf{Classification Accuracy.}
Detailed results are presented in Tab.~\ref{tab:classification_comparison}. 
The experimental outcomes demonstrate that our proposed model surpasses existing baselines across the majority of metrics, thereby establishing a new SOTA performance.
In the most challenging 40-class task, our model achieves a Top-1 accuracy of 8.27\%, significantly outperforming the runner-up, GLMNet (6.20\%). This performance advantage is maintained in the 9-class task, where our model again achieves the highest Top-1 accuracy of 22.73\%.
This consistent performance margin suggests that our model captures richer and more discriminative semantic details from the EEG signals.
Our model's capabilities are particularly evident in fine-grained attribute recognition. For the \textbf{``Human Face''} task, it achieves an accuracy of \textbf{82.47\%}, a substantial margin of over 17 percentage points above the next-best method (GLMNet at 65.10\%). 
This improvement provides robust evidence of our model's enhanced capacity for decoding specific, high-level semantic concepts.
Furthermore, our model secured leading results in other fine-grained tasks, including ``Numbers'' and ``Human''.
In summary, these results provide strong quantitative support for our model's effectiveness, verifying its ability to learn highly discriminative and semantically meaningful features for the subsequent high-fidelity video reconstruction.

\noindent\textbf{Video Reconstruction.}
We quantitatively compare our model with the EEG2Video baseline across tasks with varying class numbers (10 to 40).
Tab.~\ref{tab:reconstruction_comparison} details the evaluation, which assesses video/frame levels across semantic/pixel dimensions.
Our method significantly outperforms the EEG2Video baseline in all evaluated settings.
Semantically, our model's advantage grows with task difficulty as the number of classes increases from 10 to 40.
For instance, in the challenging 40-class setting, our model's video semantic accuracy reaches \textbf{0.284}, far surpassing the baseline's 0.159.
Similarly, our frame-based semantic accuracy (\textbf{0.241}) substantially exceeds the baseline's performance (0.138).
This result demonstrates our model provides more precise semantic guidance during the generation process.
At the pixel level, FVMD assesses video quality and temporal coherence, where lower is better.
Our model achieves an FVMD of \textbf{1637.55} in the 40-class task, markedly lower than the baseline's 2038.27.
This lower FVMD indicates superior motion smoothness and inter-frame consistency, directly validating the efficacy of our TDA module.
Furthermore, a higher SSIM indicates our generated frames possess greater structural fidelity.
In conclusion, these quantitative results verify the superiority of our dual-guidance framework across both semantic and pixel dimensions.
Our model sets a new performance benchmark for EEG-based video reconstruction in both content accuracy and visual fidelity.

\subsection{Ablation Study}

We conduct ablation studies to validate each key component of our model in Tab.~\ref{tab:ablation study}. Starting from our full model, we evaluate the contribution of each component by removing it and observing the impact on performance.

\noindent\textbf{Effect of Brain Regions.}
Among these regions, the Occipital Lobe has the most significant impact on performance, which aligns with its primary role in visual processing.
Nevertheless, removing signals from any single brain region, including the frontal, parietal, or temporal lobes, also degrades performance. 
This result underscores the collective and distributed contribution of all regional features.

\noindent\textbf{Effect of Features.}
Ablating different modal features reveals that removing the \textbf{text feature (w/o Text)} causes the most significant drop in semantic and classification performance, underscoring its critical role. The performance degradation from removing image or category features also validates the effectiveness of our multimodal fusion strategy.

\noindent\textbf{Effect of Consistency.}
Removing the structural consistency loss (\textbf{w/o Consistency}) significantly worsens the FVMD score (from 1637.55 to 1916.91). This sharp increase confirms that this objective plays a key role in enhancing the temporal coherence of the generated video.

\subsection{Visualization}

The qualitative comparison results clearly demonstrate our model's significant advantage in semantic fidelity. 
A particularly illustrative case is the reconstruction of the ``cat'' in the bottom right. 
Our model successfully generates a cat with realistic morphology and texture, exhibiting high fidelity to the ground truth. 
In stark contrast, the EEG2Video baseline fails to capture the correct semantics entirely, generating a pink, morphologically distorted object that suffers from severe content hallucination. 
This superiority is also evident in other examples: the ``car'' reconstructed by our model features a well-defined contour and accurate coloration, while the ``river'' scene is rendered with vivid colors and rich details. 
Conversely, the baseline's outputs are generally plagued by blurriness, distortion, and color deviation.

Beyond the accuracy of static content, our model also exhibits superior temporal coherence. 
In the dynamic scenes of the ``river'' and ``car,'' the videos we generate feature smooth transitions and stable morphology between frames, significantly reducing the flickering and warping artifacts common in the baseline model. 
This result empirically validates the efficacy of our \textbf{Temporal Dynamic Aligner (TDA)} in maintaining dynamic consistency. 
Furthermore, when handling complex scenes containing a ``human'' and specific ``objects'' like a guitar, our model accurately reconstructs the core scene composition and human poses, whereas the baseline struggles to produce coherent and meaningful imagery.

In summary, these visual results intuitively corroborate the superior performance of our model. 
By effectively combining rich semantic guidance with precise temporal dynamic constraints, our model is capable of generating high-quality videos that are semantically accurate, visually detailed, and temporally coherent, marking a significant improvement over existing methods.

\section{Conclusion}

This work addresses key challenges in EEG-to-video reconstruction: suboptimal use of distributed neural information, and a lack of dynamic temporal coherence. 
Therefore, we propose DynaMind, a new framework that reconstructs dynamic visual scenes by integrating multimodal semantics with temporal dynamics. 
DynaMind leverages three core modules: a Region-aware Semantic Mapper for extracting region-aware multimodal semantic priors, a Temporally-aware Dynamic Aligner for generating dynamic coherent temporal blueprint, and a Dual-guided Video Reconstructor for synthesizing the final video with both semantic and dynamic guidance. 
Experiments on the SEED-DV dataset show that DynaMind achieves state-of-the-art results, outperforming existing methods in both direct EEG classification and reconstructed video quality with notable improvements in semantic accuracy, pixel fidelity, and temporal consistency. 
Ablation studies confirm the necessity of each component, signifying a major advance in decoding visual experiences from non-invasive brain recordings.

\bibliography{aaai2026}

\clearpage
\begin{appendix}

\section*{Appendix}
This appendix provides additional details, analyses, and experiments to support our main findings. The contents are organized as follows: Appendix~\ref{sec:exp} provides detailed descriptions of our experimental settings; Appendix~\ref{sec:eval} provides detailed descriptions of our evaluation metrics; Appendix~\ref{sec:baseline} provides detailed descriptions about our compared baselines; Appendix~\ref{sec:vis} provides extra visualization about successful and failure cases.

\section{Experimental Setup}\label{sec:exp}
\subsection{Dataset}
We use the public SEED-DV dataset~\cite{liu2024eeg2video}, which comprises 40 distinct visual concepts.
These concepts fall into nine higher-level categories, including Land Animal, Water Animal, Plant, Exercise, Human, Natural Scene, Food, Musical Instrument, and Transportation.
For each concept, the dataset provides 35 video clips along with the corresponding EEG signals recorded from 20 subjects.

\subsection{Implementation Details}
Our model comprises three core modules, each with a distinct implementation and training protocol:
\begin{itemize}
    \item \textbf{Regional-aware Semantic Mapper~(RSM):} EEG data is parcellated into $K=4$ neuroscientifically-informed brain regions. Each region's data is processed by independent layers to produce 512-dimensional embeddings, which are fused and projected into a unified 1024-dimensional feature. This feature is aligned with the text and image feature space of a pre-trained CLIP ViT-L/14 model. The RSM encoder and alignment mappings are jointly trained for 300 epochs (learning rate $1 \times 10^{-5}$). A separate DALLE-inspired prior module is then trained for 1000 epochs (learning rate $2 \times 10^{-5}$).
    \item \textbf{Temporal-aware Dynamic Aligner~(TDA):} EEG signals are segmented into 6-step windows. An encoder processes these windows to generate a temporal feature sequence. The module is trained for 300 epochs (learning rate $1 \times 10^{-5}$) to align its output with VAE latent features derived from the ground-truth video via DDIM inversion.
    \item \textbf{Dual-Guidance Video Reconstructor~(DGVR):} We fine-tune a Stable Diffusion v1.4 model for 200 epochs with a learning rate of $3 \times 10^{-5}$, following the protocol in~\cite{wu2023tune}. Training data consists of 6-frame video clips at a $512 \times 288$ resolution and 3 fps. All models are trained on an NVIDIA A800 GPU using a cosine annealing learning rate scheduler.
\end{itemize}

\section{Evaluation Metrics}\label{sec:eval}
Our evaluation follows the protocol established by EEG2Video~\cite{liu2024eeg2video} to ensure a fair comparison.
First, we assess frame-level quality via two metrics: the Structural Similarity Index~(SSIM)~\cite{wang2004image} for pixel fidelity, and N-way top-K accuracy using a CLIP-based classifier~\cite{radford2021learning} for semantic content.
Acknowledging that static frame analysis cannot capture temporal dynamics, we introduce complementary video-level metrics.
We employ a VideoMAE~\cite{tong2022videomae} classifier, pre-trained on the Kinetics-400 dataset~\cite{kay2017kinetics}, to evaluate the dynamic semantic coherence of the generated video clips.
Furthermore, we use the Frame-wise Video Motion Distance~(FVMD)~\cite{liu2024fr} metric to assess the temporal consistency and motion smoothness of the reconstructed videos.

\subsubsection{Structural Similarity Index (SSIM)}
SSIM assesses the pixel-level fidelity of a reconstructed frame by comparing its structural information with a ground-truth frame. It is a perception-based model that considers changes in luminance, contrast, and structure. For two image windows $x$ (ground truth) and $y$ (reconstruction), SSIM is defined as:
\begin{equation}
    \text{SSIM}(x, y) = \frac{(2\mu_x\mu_y + C_1)(2\sigma_{xy} + C_2)}{(\mu_x^2 + \mu_y^2 + C_1)(\sigma_x^2 + \sigma_y^2 + C_2)}.
\end{equation}
where $\mu_x$ and $\mu_y$ are the average pixel values, $\sigma_x^2$ and $\sigma_y^2$ are the variances, and $\sigma_{xy}$ is the covariance of $x$ and $y$. $C_1 = (k_1L)^2$ and $C_2 = (k_2L)^2$ are stabilization constants, where $L$ is the dynamic range of pixel values. The final score is the mean SSIM over all windows.

\subsubsection{Frame-wise Video Motion Distance (FVMD)}
FVMD evaluates the temporal consistency and motion realism among frames. It is analogous to Fréchet Inception Distance but is applied to motion features extracted from consecutive frames. First, motion features (\textit{e.g.}, from optical flow) are extracted for all pairs of adjacent frames from both ground-truth videos and generated videos. These sets of features are then modeled as multivariate Gaussian distributions. FVMD is the Fréchet distance between these two distributions:
\begin{equation}
    \text{FVMD} = \|\mu_r - \mu_g\|^2_2 + \text{Tr}\left(\Sigma_r + \Sigma_g - 2(\Sigma_r\Sigma_g)^{1/2}\right).
\end{equation}
where $(\mu_r, \Sigma_r)$ and $(\mu_g, \Sigma_g)$ are the mean and covariance of the motion features from the real and generated videos, respectively, and $\text{Tr}(\cdot)$ is the trace of a matrix. A lower FVMD score indicates more realistic and coherent motion.

\begin{figure*}[t]
 \centering
 \includegraphics[width=0.92\textwidth]{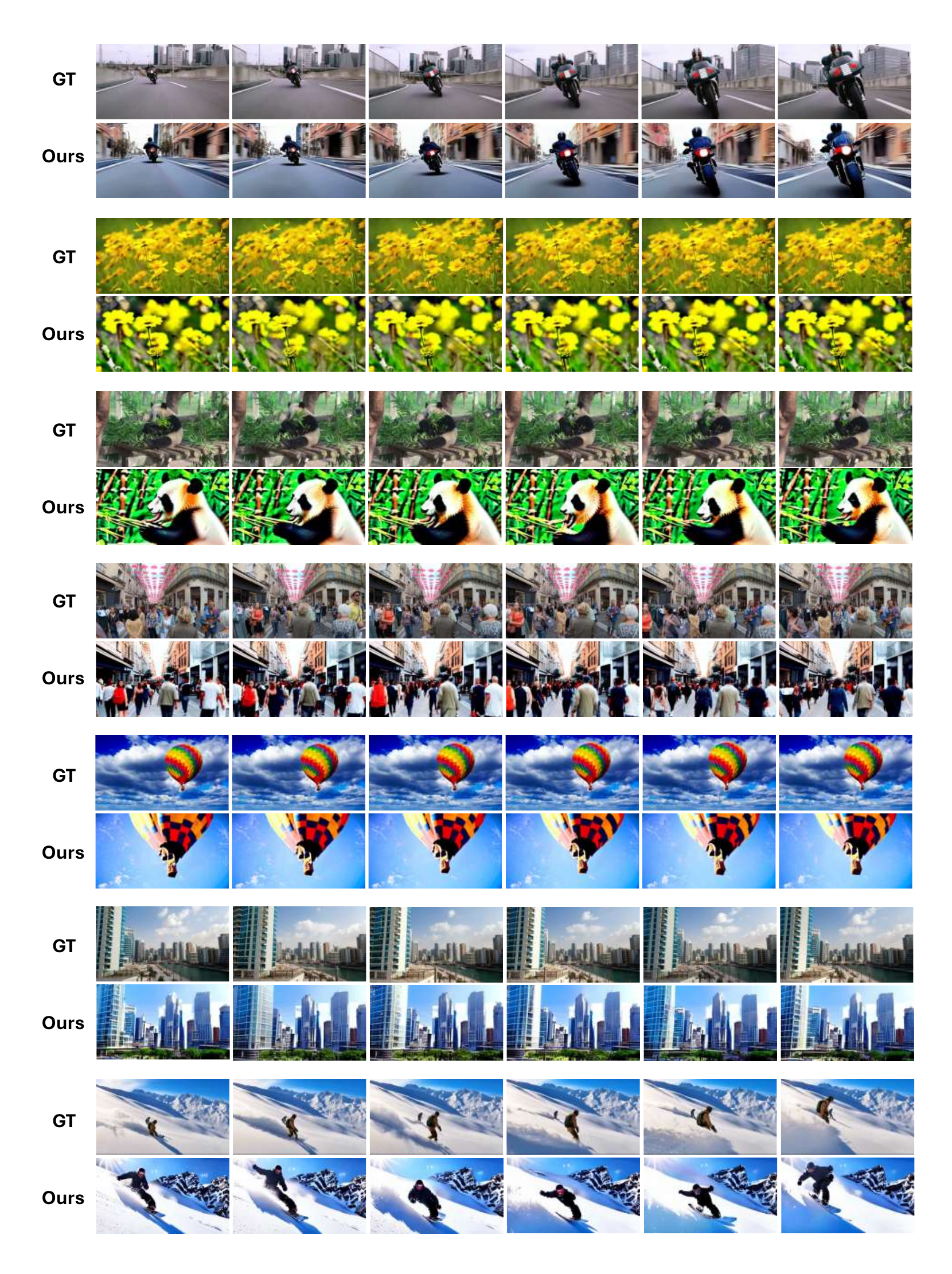} 
 \caption{Various videos reconstruction samples.}
 \label{fig:more1}
\end{figure*}

\begin{figure*}[t]
 \centering
 \includegraphics[width=0.92\textwidth]{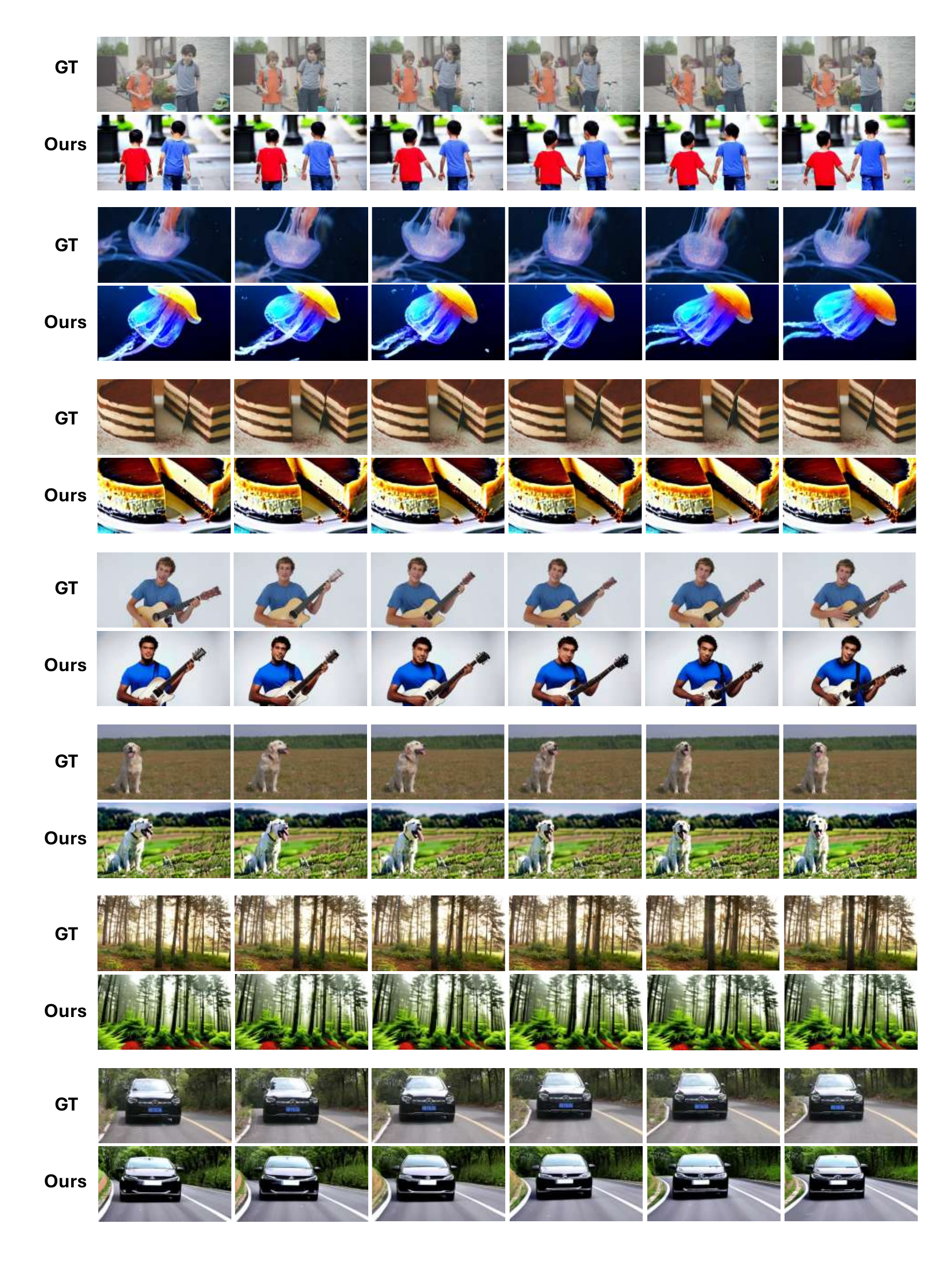} 
 \caption{Various videos reconstruction samples.}
 \label{fig:more2}
\end{figure*}

\begin{figure*}[t]
 \centering
 \includegraphics[width=0.92\textwidth]{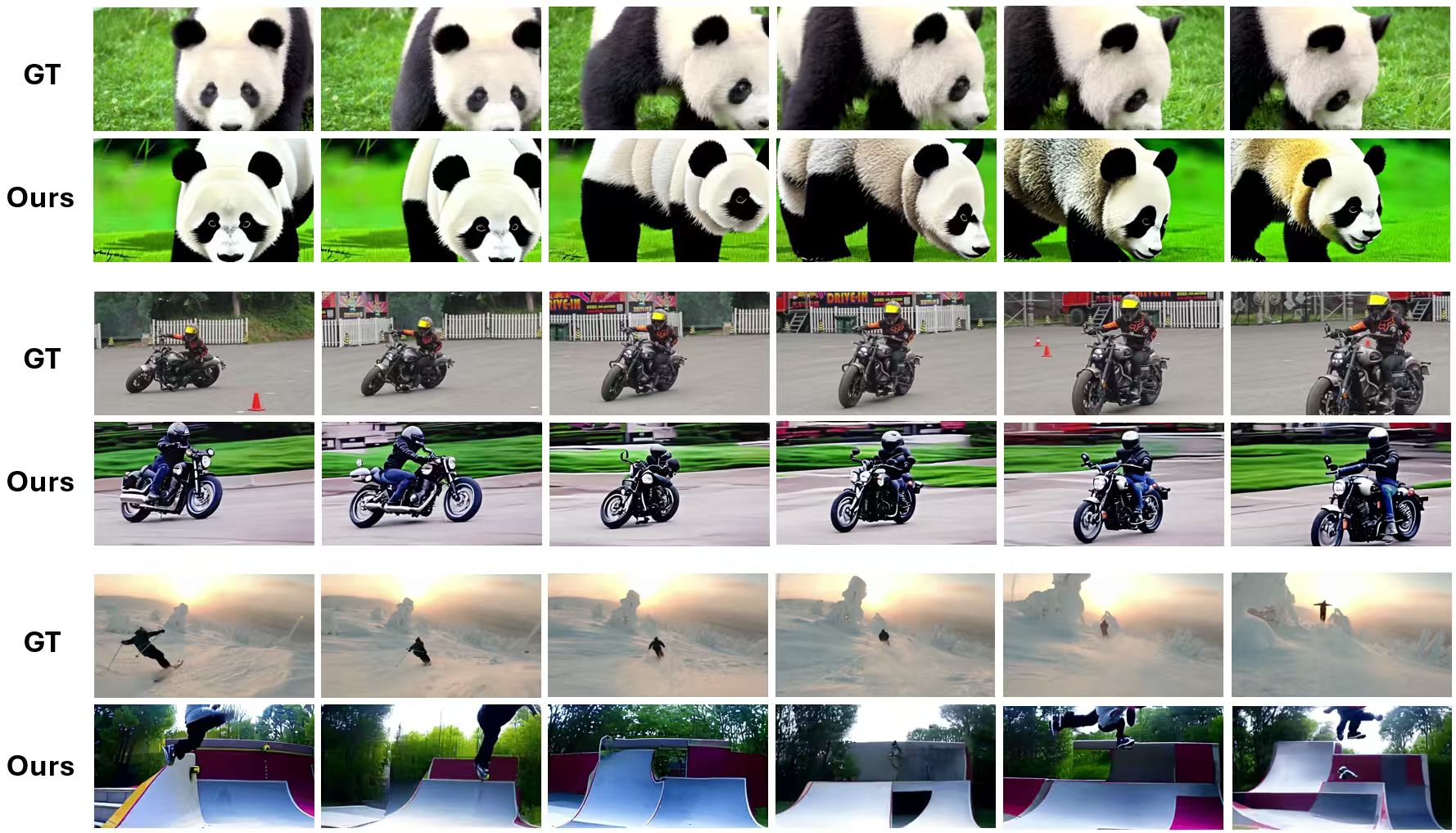} 
 \caption{Some failure samples.}
 \label{fig:failure}
\end{figure*}

\subsubsection{N-way top-K Accuracy on Frames and Videos}
This metric evaluates the semantic accuracy of a single reconstructed frame or whole video. 
For a given frame or video, its class is predicted from a set of $N$ randomly chosen classes (the ground-truth class plus $N-1$ ``distractor'' classes). The prediction is considered correct if the ground-truth class is within the top-$K$ predicted probabilities. For a test set of size $M$, the accuracy is:
\begin{equation}
    \text{Acc}_{N,K} = \frac{1}{M} \sum_{i=1}^{M} \mathbb{I}(y_i \in \text{TopK}(\hat{P}_i(S_N))).
\end{equation}
where $y_i$ is the ground-truth label for sample $i$, $S_N$ is the set of $N$ classes, $\hat{P}_i(S_N)$ are the predicted probabilities for classes in $S_N$, $\text{TopK}(\cdot)$ returns the set of $K$ classes with the highest probabilities, and $\mathbb{I}(\cdot)$ is the indicator function.
For both frame and video-level evaluations, we employ 40-way and 2-way top-1 accuracy to assess the semantic quality of the reconstructed content.
For the frame-level classifier, we follow the protocol in~\citet{liu2024eeg2video} and fine-tune a pre-trained CLIP encoder on the ground-truth frames for the 40-way visual concept classification task.
Similarly, for the video-level classifier, a pre-trained VideoMAE model is fine-tuned on the ground-truth video clips, consistent with the same evaluation protocol~\cite{liu2024eeg2video}.

\section{Baselines}\label{sec:baseline}
We evaluate our model's performance against a comprehensive set of strong and representative baselines from the field of EEG analysis.
These selected baselines range from foundational CNN architectures to recent state-of-the-art models.

\noindent\textbf{ShallowNet and DeepNet} are two foundational CNN architectures for EEG decoding proposed by~\citet{schirrmeister2017deep}. 
\textbf{ShallowNet} employs both temporal and spatial convolutional layers, an architecture effective for decoding frequency power information.
In contrast, \textbf{DeepNet} is a deeper, VGG-inspired network designed to learn complex features through stacked convolutional blocks.

\noindent\textbf{EEGNet}, from~\citet{lawhern2018eegnet}, is a compact and efficient CNN architecture designed for EEG-based brain-computer interfaces (BCIs).
The model uses depthwise and separable convolutions to extract temporal and spatial features with fewer parameters, which enhances its generalizability across different BCI paradigms.

\noindent\textbf{Conformer}, from~\citet{song2022eeg}, is an architecture for EEG signal processing that combines the strengths of both CNNs and Transformers.
The model employs convolutional layers to extract local features and a multi-head self-attention mechanism to model global correlations, an approach proven effective for various EEG decoding tasks.

\noindent\textbf{TSConv.}
The Temporal-Spatial Convolutional network~(TSConv), from~\citet{song2023decoding}, is an advanced CNN-based model for EEG analysis. 
It uses dedicated convolutional blocks to explicitly learn features along both temporal and spatial dimensions, which enhances its ability to extract discriminative information.

\noindent\textbf{GLMNet}, from EEG2Video~\citet{liu2024eeg2video}, is a recent state-of-the-art method that serves as a primary baseline for our work. 
This sophisticated network extracts high-level semantic features from EEG for visual decoding, representing the key performance benchmark we aim to surpass.

\section{More Visualizations}\label{sec:vis}

\subsubsection{More Successful Examples of Reconstruction}
This section provides additional visualization results to further demonstrate the robustness and generality of our proposed model.
As shown in Figures~\ref{fig:more1} and~\ref{fig:more2}, each example directly compares our model's reconstruction (\textbf{Ours}) with the ground truth video (\textbf{GT}).
These supplementary examples cover a diverse range of semantic categories, including:
\begin{itemize}
    \item \textbf{Human-related Scenarios:} human scenes,musical activities, etc.
    \item \textbf{Nature-related Scenarios:}  land animals, water animals, Plants, landscape, etc.
    \item \textbf{Man-made Objects:} Food, Transportation, Cities, etc.
\end{itemize}
Across these varied scenarios, our model consistently generates reconstructions that are highly faithful to the ground truth videos in semantic content, color, and temporal coherence. 
This visual evidence, combined with the quantitative results in the main paper, validates our model's robust capability for decoding complex dynamic visual experiences from non-invasive EEG signals.

\subsubsection{Some failure samples}

Although our model performs well in most cases, analyzing its failure cases allows us to more clearly identify its limitations. As shown in Fig.~\ref{fig:failure}, these limitations primarily fall into three categories:
\begin{itemize}
    \item \textbf{Partial Frame Distortion:} In some cases, while the core object is correctly identified, parts of the frame or background still exhibit visual distortion. For example, in the ``panda'' scene, the reconstruction captured the basic shape of the panda, but as the perspective changes in the video, the subject undergoes morphological distortion in some frames.
    \item \textbf{Incoherent Object Motion:} In the "motorcycle" example, although the model generates semantically correct single-frame images, there are issues with inconsistency or mismatch with the real world in terms of the object's direction of motion and movement pattern.
    \item \textbf{Incorrect Semantics:} In the ``skiing'' example, the model misinterprets a person on a blurry snow slope as someone skateboarding on a ramp. This suggests that for visually ambiguous inputs, the model can still converge to an incorrect semantic concept.
\end{itemize}
\end{appendix}

\end{document}